\pdfoutput=1

\documentclass[11pt]{article}

\usepackage[]{EMNLP2023}

\usepackage{times}

\usepackage{latexsym}
\usepackage{booktabs}
\usepackage{arydshln}
\usepackage{colortbl}
\newcommand*\rot{\rotatebox{90}}

\usepackage{multirow}
\usepackage{xcolor,pifont}
\newcommand*\colourcheck[1]{%
  \expandafter\newcommand\csname #1check\endcsname{\textcolor{#1}{\ding{52}}}%
}
\colourcheck{blue}
\colourcheck{green}
\colourcheck{red}
\colourcheck{black}

\usepackage{highlight}
\newcolumntype{a}{>{\columncolor{Gray}}c}

\usepackage[nointegrals]{wasysym}
\definecolor{darkseagreen}{rgb}{0.46, 0.74, 0.46}
\definecolor{alizarin}{rgb}{0.82, 0.1, 0.26}
\usepackage{CJKutf8}
\newcommand{\zh}[1]{\begin{CJK}{UTF8}{gbsn}#1\end{CJK}}
\usepackage{amssymb}
\usepackage{pifont}
\newcommand{\xmark}{\ding{55}}%

\definecolor{CustomBlue}{RGB}{57,83,191}
\usepackage[most]{tcolorbox}
\newtcbox{\clustertab}[1]{on line, box align=base, colback={#1},colframe={#1},size=fbox,arc=2pt,top=-1.5pt, bottom=-1.5pt, left=-1.5pt, right=-1.5pt, boxrule=0pt, enlarge left by=1pt}

\newcommand{\goodS}[1]{{\small\clustertab{set10-blue!5}{\color{blue!80}$\downarrow  \mathbf{#1}$}}}
\newcommand{\goodchrfS}[1]{{\small\clustertab{set10-blue!5}{\color{blue!80}$\uparrow  \mathbf{#1}$}}}

\newcommand{\goodM}[1]{{\small\clustertab{set10-blue!10}{\color{blue!80}$\downarrow  \mathbf{#1}$}}}

\newcommand{\goodL}[1]{{\small\clustertab{set10-blue!15}{\color{blue!80}$\downarrow  \mathbf{#1}$}}}
\newcommand{\goodXL}[1]{{\small\clustertab{set10-blue!30}{\color{blue!80}$\downarrow  \mathbf{#1}$}}}

\newcommand{\bad}[1]{{\small\clustertab{google-red!10}{\color{google-red}{$\uparrow \mathbf{#1}$}}}}
\newcommand{\badchrf}[1]{{\small\clustertab{google-red!20}{\color{google-red}{$\downarrow \mathbf{#1}$}}}}
\newcommand{\badchrfS}[1]{{\small\clustertab{google-red!5}{\color{google-red}{$\downarrow \mathbf{#1}$}}}}

\newcommand{\nosig}[1]{{\small\clustertab{yellow!15}{\color{orange}$\downarrow  \mathbf{#1}$}}}
\newcommand{\sigC}[1]{{\small\clustertab{darkseagreen!15}{\color{darkseagreen}$\downarrow  \mathbf{#1}$}}}
\newcommand{\sigB}[1]{{\small\clustertab{darkseagreen!15}{\color{darkseagreen}$\downarrow  \mathbf{#1}$}}}
\newcommand{\sigA}[1]{{\small\clustertab{blue!10}{\color{blue!80}$\downarrow  \mathbf{#1}$}}}

\newcommand{\clustersig}[1]{{\small\clustertab{orange!80}{\color{orange!10}$ \mathbf{#1}$}}}
\newcommand{\clustersigde}[1]{{\small\clustertab{darkseagreen}{\color{darkseagreen!10}$ \mathbf{#1}$}}}
\newcommand{\clustersigzh}[1]{{\small\clustertab{blue!50}{\color{blue!10}$ \mathbf{#1}$}}}

\definecolor{Gray}{gray}{0.85}

\usepackage{colortbl}

\newcolumntype{a}{>{\columncolor{Gray}}c}
\definecolor{set10-blue}{HTML}{4169E1}
\definecolor{google-red}{HTML}{de5246}

\newcommand{\contentOmitted}{\textit{[content omitted]}}

\usepackage[T1]{fontenc}

\usepackage[utf8]{inputenc}
\usepackage{subcaption}

\usepackage{microtype}

\usepackage{inconsolata}
\usepackage{graphicx}
\usepackage{xcolor}
\usepackage{xcolor}

\newcommand{\ignore}[1]{}
\title{Translating Step-by-Step: Decomposing the Translation Process for Improved Translation Quality of Long-Form Texts}

\author{Eleftheria Briakou, Jiaming Luo,  Colin Cherry, Markus Freitag \\
  Google
 \\
 \texttt{\{ebriakou,jmluo,colincherry,freitag\}@google.com}
 } 

\begin{document}
\maketitle
\begin{abstract}
In this paper we present a step-by-step approach to long-form text translation, drawing on established processes in translation studies. Instead of viewing machine translation as a single, monolithic task, we propose a framework that engages language models in a multi-turn interaction, encompassing pre-translation research, drafting, refining, and proofreading, resulting in progressively improved translations.
Extensive automatic evaluations using Gemini $1.5$ Pro across ten language pairs show
that translating step-by-step yields large translation quality improvements over conventional zero-shot prompting approaches and earlier human-like baseline strategies, resulting in state-of-the-art results on \textsc{wmt} $2024$.
\end{abstract}

\section{Introduction}

Machine Translation (\textsc{mt}) has been traditionally seen as a sequence transduction task that maps a source text from one language to an equivalent translation in another language.
While this simplified definition of the task served the modeling capabilities of statistical and neural machine translation systems for many years, recent advancements in large language modeling offer promise for re-defining \textsc{mt} to align more closely with human translation processes. This shift prompts us back to a fundamental question: \textit{what does a good translation process look like}?

Thankfully, this question has been a long-debated topic in the field of translation studies. Despite the lack of consensus around the nature of cognitive steps involved when humans translate, a common thread is apparent, i.e.,  translation is a multi-faceted \textit{process} encompassing several sub-tasks that navigate a bilingual landscape.
This view of translation finds a parallel in the rise of the ``chain-of-thought'' paradigm popularized by large language models (\textsc{llm})~\cite{cot}. That is, instead of attempting to generate the response to a complex task directly, \textsc{llm}s are prompted to derive their final answer by decomposing the original task into several simpler sub-tasks.

\begin{figure}[!t]
    \centering
    \includegraphics[scale=0.63]{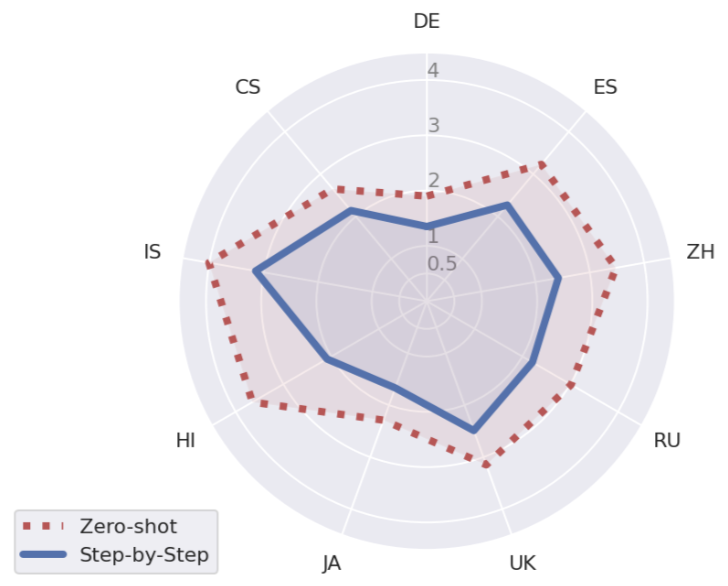}
    \caption{MetricX-$23$ quality improvements (where lower scores indicate better translation quality) on document-level translation on the \textsc{wmt}$24$ test set. Translate step-by-step with Gemini $1.5$ Pro consistently outperforms zero-shot translation.}
    \label{fig:sbys_main_fig_gains}
\end{figure}

But, what form would chain-of-thought take in the context of \textsc{mt}? 
While initial attempts to model the entire translation process using complex multi-stage processes has shown mixed results~\cite{Wu2024PerhapsBH},
explicitly modeling certain pre-translation or post-translation processes has led to more consistent gains in translation quality. On the pre-translation side,  \citet{He2023ExploringHT} proposes to generate multiple translation candidates conditioned on self-generated translation-related knowledge. On the post-translation side, recent research threads prompt \textsc{llm}s for refinement with \cite{feng2024improving, Xu2023LLMRefinePA, Ki2024GuidingLL} or without~\cite{Chen2023IterativeTR} external quality estimation feedback.

Despite the promising results reported by prior work on decomposing and re-ranking \textsc{mt} with \textsc{llm}s, 
it still remains unclear whether \textsc{llm}s can benefit from modeling the \textit{entire spectrum of translation processes}. In this work, drawing on literature from translation studies, we view \textsc{mt} as a complex and iterative task adhering to distinct steps, i.e., pre-translation research, drafting, refining, and proofreading. Based on this framework, we ask: \textit{How well can \textsc{llm}s translate in a step-by-step manner that draws from translation processes}?

Taking Gemini $1.5$ Pro~\cite{Reid2024Gemini1U} as a case study, we start by designing instruction prompts for various translation subtasks. Concretely, our framework implements a \textit{\textbf{multi-turn}} interaction with Gemini that breaks down the translation process into four distinct stages. It begins by prompting the model to conduct background research that identifies potential challenges in translating the source text (\textit{\textbf{research}} phase). The next interaction focuses on drafting an initial translation prioritizing faithfulness to the source text (\textit{\textbf{drafting}} phase). This draft is then revised in subsequent turns, ensuring a polished final translation (\textit{\textbf{refinement}} and \textit{\textbf{proofreading}} phases). 

To align better with human translation processes, we test the \textit{translate step-by-step} framework on long-form documents derived from the general \textsc{mt} shared tasks for \textsc{wmt} $2023$~\cite{kocmi-etal-2023-findings} and \textsc{wmt} $2024$. We evaluate out-of-English translation for ten languages, namely Chinese (\textsc{zh}), 
Ukrainian (\textsc{uk}),
Russian (\textsc{ru}),
Japanese (\textsc{ja}),
Hebrew (\textsc{he}),
Czech (\textsc{cs}),
German (\textsc{de}),
Hindi (\textsc{hi}), 
Icelandic (\textsc{is}),
and
Spanish (\textsc{es}).
Extensive automatic evaluation according to both reference-based and 
\textsc{qe}-based versions of MetricX-$23$~\cite{juraska-etal-2023-metricx} show that translating step-by-step yields strong translation quality improvements across all languages and test sets studied (see Figure~\ref{fig:sbys_main_fig_gains}).  

\section{Background}

With the recent rise of \textsc{llm}s, machine translation is going through a gradual but significant paradigm shift. While much research is focusing on how \textsc{llm}s' training data are improving their \textsc{mt} capabilities~\cite{Xu2023APS, Alves2024TowerAO},
there are also many opportunities to improve how existing \textsc{llm}s can be best used for translation.
This becomes evident in recent research that explores ways to augment and refine \textsc{mt} to align better with human translation processes.
To navigate the diverse landscape of \textsc{llm}-driven research, we summarize key studies in Table~\ref{tab:prior_work} along their four most distinct dimensions:

\begin{table}[!t]
    \scalebox{0.87}{
    \centering
    \begin{tabular}{l>{\columncolor[gray]{0.9}}c>{\columncolor[gray]{0.9}}c>{\columncolor[gray]{0.95}}c>{\columncolor[gray]{0.90}}c>{\columncolor[gray]{0.95}}l} 
    \textsc{paper} & \rot{\textsc{pre-tr.}}  & \rot{\textsc{post-tr.}} & \rot{\textsc{dev.}}  & \rot{\textsc{param.}} & \rot{\textsc{steps}}\\
    \citet{He2023ExploringHT} & \bluecheck    & \xmark & \xmark  & \xmark & $3$-$4$ \\
    \citet{Xu2023LLMRefinePA} & \xmark & \bluecheck & \bluecheck  & \xmark & Iterative\\ 
    \citet{feng2024improving} & \xmark &  \bluecheck & \xmark & \xmark  & $3$\\
    \citet{Huang2024AligningTU} & \xmark &  \bluecheck & \bluecheck  & \xmark & $3$\\
    \citet{Li_Chen_Yuan_Wu_Yang_Tao_Xiao_2024} & \bluecheck & \xmark  & \xmark  & \xmark & $1$\\
    \citet{Chen2023IterativeTR} & \xmark & \bluecheck & \xmark & \bluecheck  & Iterative\\
    \citet{Ki2024GuidingLL} & \xmark & \bluecheck & \xmark & \xmark & $1$\\
    \citet{Wu2024PerhapsBH} & \bluecheck & \bluecheck & \xmark  & \bluecheck & Iterative\\
    Step-by-Step (ours) & \bluecheck & \bluecheck & \bluecheck  & \bluecheck & $4$ \\
    \\

    \end{tabular}}
    \caption{List of prior work leveraging \textsc{llm}s to improve translation quality by modeling either pre- or post-translation processes (\textsc{pre-tr.} or \textsc{post-tr.}). For each study we also note key aspects of their methodology:
    whether prompting strategies are developed on a separate development set (\textsc{dev.}), whether the approach relies solely on the \textsc{llm}'s parametric knowledge (\textsc{param.}), and the number of steps in the pipeline.}
    \label{tab:prior_work}
\end{table}
\begin{figure*}[!ht]
    \centering
    \includegraphics[scale=0.3]{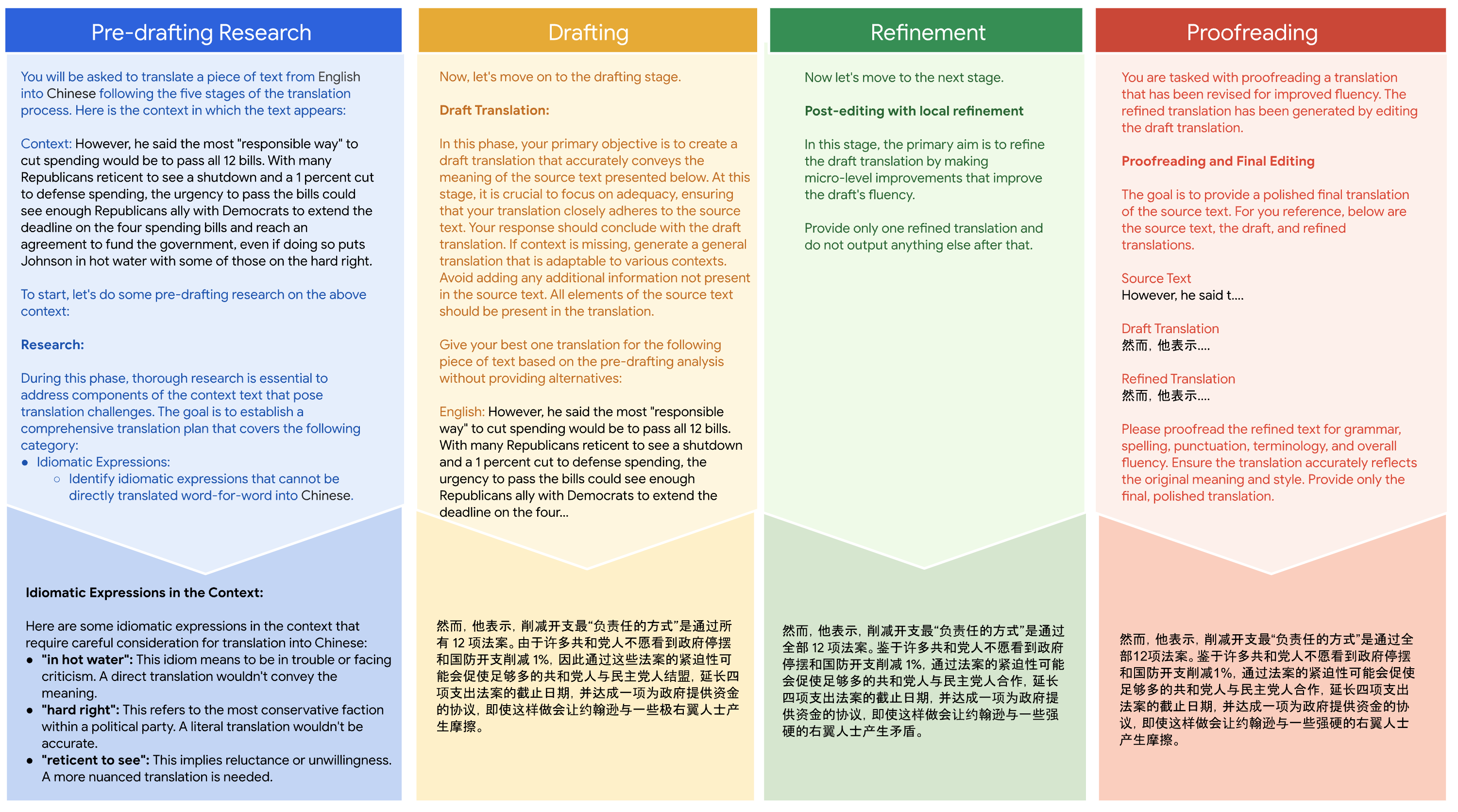}
    \caption{Translate Step-by-Step prompting framework. User prompts (top) and Gemini's responses (bottom) for the translation of an English document into Chinese. The full prompts for each step also appear in \S\ref{sec:appendix_prompts}.}
    \label{fig:sbys_example}
\end{figure*}

\begin{itemize}
    \item \textbf{Temporal Focus}: This differentiating factor is based on whether an \textsc{llm} is engaged in the translation process before (pre-translation) or after (post-translation) an initial translation is produced (whether by the same \textsc{llm} or a different system).
    \item \textbf{Parametric~vs.\ External ~Knowledge}: This dimension focuses on whether \textsc{llm}s rely solely on their internal, learned knowledge (encoded in their parameters) or whether they use external resources, i.e., dictionaries, knowledge bases, retrieval engines or \textsc{qe}-based metrics~\cite{mallen-etal-2023-trust}.
    \item \textbf{Reported Prompt Development}: This dimension considers whether the prompting strategies are clearly developed on separate development sets, as reported in papers.\footnote{We include this column not to cast aspersions on previous work, but to  encourage a culture moving forward where prompt-based research uses and reports a development set. From personal communication, some of the works receiving an ``\xmark'' here underwent little to no prompt optimization.}
    \item \textbf{Number of Steps}: This dimension counts the number of distinct steps that are used in multi-turn interactions with the \textsc{llm}.
\end{itemize}

Table~\ref{tab:prior_work} shows a clear trend: most studies focus on post-translation refinement. These approaches predominantly rely on external feedback to identify and correct errors, using either automatic metrics~\cite{feng2024improving, Xu2023LLMRefinePA, Huang2024AligningTU} or human annotations of translation errors~\cite{Ki2024GuidingLL}. A notable exception is the study of \citet{Chen2023IterativeTR}, which shows that \textsc{llm}s can iteratively refine their own outputs using only their parametric knowledge. 

Comparatively fewer studies explore the pre-translation stage, investigating how \textsc{llm}s can utilize background information to enhance their translation quality. \citet{He2023ExploringHT} explores this by prompting \textsc{llm}s for  
different types of background information (similar examples, topics and keywords) related to the source text.
However, they find that this knowledge alone is insufficient to improve the model's translation quality, and ultimately rely on external \textsc{qe} feedback for selection. In contrast, \citet{Li_Chen_Yuan_Wu_Yang_Tao_Xiao_2024} operationalizes background research by incorporating idiom definitions retrieved from an external knowledge base. 

A notable exception to the above is the recent work of \citet{Wu2024PerhapsBH} which, similar to our approach, explores modeling the entire spectrum of translation processes. While conceptually aligned with our step-by-step approach, their framework is significantly more complex, with $30$ distinct \textsc{llm} roles interacting iteratively. Their use of non-standard metrics makes it difficult to gauge the method's success: the human evaluation does not give annotators source or reference texts, while the bilingual automatic evaluation collects only preference decisions using the same model family as the method being tested.

Overall, in contrast to prior work, which often relies on complex multi-stage processes and external resources, 
our goal is to streamline the translation process, \textbf{\textit{unifying pre- and post-translation stages within one framework, by accessing only the \textsc{llm}'s parametric knowledge throughout}}. We emphasize the methodological soundness of our pipeline by developing it on a separate development set, a practice not yet standardized in this area.

\section{Translate Step-by-Step}

Drawing on existing literature on translation studies~\cite{Borg2018ThePO}, we design a series of staged prompts that attempt to map the translation process to instructions.
This approach views translation as a multi-turn interaction with an \textsc{llm} where each prompt guides the model's next action.
Below, we describe those stages, along with what their function in the translation process is and how they are operationalized as  instruction-following tasks. These stages are further illustrated in Figure~\ref{fig:sbys_example}.

\paragraph{Pre-translation Research} Mirroring the human translation processes, our framework incorporates a pre-translation research stage. This stage primarily focuses on using the source text~\cite{Mossop2000TheWP} to identify potential translation challenges drawing on real-world knowledge and knowledge of the target language~\cite{Dimitrova2005ExpertiseAE}.  We model this stage by prompting the \textsc{llm} to identify and explain phrases of the source text that cannot be translated word-for-word into the target language.

\paragraph{Drafting} \label{sec:method_draft} Following the pre-translation research, the next stage aims at producing a draft translation, i.e., ``the first stab at the rewriting''  \cite{peterbush}. This stage represents an initial attempt at rendering the source text into the target language. To that end, we initiate a subsequent interaction and prompt the model to focus on adequacy at this stage, ensuring the draft faithfully captures the meaning of the source.

\paragraph{Refinement} The \textit{post-drafting} stages are defined as editing tasks, with the goal of improving the overall quality of the draft translation. We define the first post-drafting stage as a subsequent interaction where the \textsc{llm} is prompted to improve the draft's fluency such that the text works on its own~\cite{Borg2018ThePO}.

\paragraph{Proofreading} At the final, post-drafting stage, we task the \textsc{llm} with the role of proofreading the refined translation to ensure it delivers a polished translation. We model this stage as a new conversation with the \textsc{llm}, rather than a subsequent interaction, drawing inspiration from human studies suggesting that proofreading requires a new perspective after a break from revising~\cite{Shih2013TranslatorsEP}.

\subsection{Lessons During Development}

While developing the above method, we found two factors to be important for the success of this approach: working at the document level and representing multi-step interactions as conversations.

\paragraph{Working at the Document Level} Our multi-step process became more effective as we moved from the segments provided by \textsc{wmt} to working on multi-segment documents (see \S\ref{sec:doc_setting} for details on the setup). This had a large effect on the pre-translation research step, changing it in two ways. First, some phrases that appeared idiomatic or difficult at the segment level disappeared, as their translations became clear with context. Second, the \textsc{llm} began identifying larger phrases.  The refinement step also improved according to automatic metrics. We verified that our shift to the document level was either neutral or an improvement for our baselines (\S\ref{sec:prior_work}).

\paragraph{Multi-step Interactions as Conversations} Modern \textsc{llm}s use special markers to indicate human versus assistant turns in multi-turn interactions. When building an automated process like translate step-by-step, for each step, one has the option to either use previous outputs to build a completely new query that summarizes all previous interactions, or to continue the conversation, allowing the \textsc{llm} to see all previous steps with its own outputs clearly marked. With the exception of the proofreading step, we found that continuing the conversation improved performance. Also, breaking the conversation into smaller turns helps with modularity for ablations.

\section{Experimental Setting}\label{sec:doc_setting}

We start by evaluating the translate step-by-step approach on the task of document-level translation. The experimental setting is described below.

\paragraph{Model Settings} Throughout our experiments we use Gemini $1.5$ Pro. All model outputs are generated with greedy decoding. All model prompts are provided in Appendix~\ref{sec:appendix_prompts}. In zero-shot mode, the model is instructed to translate the source text directly, without providing any explanations. 

To effectively isolate the artifacts from pre-translation research, we employ a secondary model call. This call restructures the natural language output into a \textsc{json} object, simplifying the parsing process for extracting artifacts.
\begin{table}[!t]
    \scalebox{0.86}{
    \centering
    \begin{tabular}{lrrrr}
    \rowcolor{gray!20}
    Domain            & Literary & News & Social & Speech \\ 
    \# Docs.        &  $40$   & $43$ & $48$ & $111$ \\
    Avg. Length  & $192$    & $184$ & $164$ & $73$ \\
    \end{tabular}}
    \caption{Per-domain statistics for \textsc{wmt} $2024$.}
    \label{tab:per_domain_wmt24_stats}
\end{table}

\begin{table*}[!ht]
    \centering
    \scalebox{0.75}{
    \begin{tabular}{rlllllllllll>{\columncolor[gray]{0.9}}c}
    \arrayrulecolor{gray!20}
    \rowcolor{blue!10}
     & \rot{\textit{Research}} & \rot{\textit{Draft}} & \rot{\textit{Refinement}} & \rot{\textit{Proofreading}} & \textbf{\textsc{zh}} & \textbf{\textsc{uk}} & \textbf{\textsc{ru}} & \textbf{\textsc{ja}} & \textbf{\textsc{he}} & \textbf{\textsc{cs}} & \textbf{\textsc{de}} & \textbf{\textsc{average}} \\
    \multicolumn{13}{l}{\textit{Ref-based}} \\

    $1.$ & \Circle & \Circle & \Circle & \Circle & $3.64$ & $4.18$ & $3.32$ & $2.59$ & $4.36$ & $2.82$ & $1.82$  &  \abref{3.25} \\
    $2.$ & \Circle & \CIRCLE & \Circle & \Circle & $3.48$ \nosig{0.16} & $4.16$ \nosig{0.02} & $3.32$ \nosig{0.00} & $2.47$ \nosig{0.12} & $4.54$ \bad{0.18} & $2.67$ \nosig{0.15} & $1.92$ \bad{0.10}  &  \abref{3.22} \\
    $3.$ & \Circle & \Circle & \CIRCLE & \Circle & $2.92$ \sigA{0.72} & $3.32$ \sigA{0.86} & $2.43$ \sigA{0.89} & $2.19$ \sigC{0.40} & $3.24$ \sigA{1.12} & $2.35$ \sigA{0.47} & $1.31$ \sigA{0.51}  & \abref{2.54} \\
    $4.$ & \Circle & \CIRCLE & \CIRCLE & \Circle & $2.85$ \sigA{0.79} & $3.06$ \sigA{1.12} & $2.54$ \sigB{0.78} & $2.09$ \sigA{0.50} & $3.18$ \sigA{1.18} & $2.22$ \sigB{0.60} & $1.37$ \sigA{0.45}  &  \abref{2.47} \\
    $5.$ & \CIRCLE & \CIRCLE & \Circle & \Circle & $3.00$ \sigA{0.64} & $3.46$ \sigA{0.72} & $2.56$ \sigA{0.76} & $2.05$ \sigA{0.53} & $3.89$ \nosig{0.47} & $1.97$ \sigA{0.85} & $1.56$ \sigA{0.26}  &  \abref{2.64} \\
    $6.$ & \CIRCLE & \CIRCLE & \CIRCLE & \Circle & $2.63$ \sigA{1.01} & $2.70$ \sigA{1.47} & $2.13$ \sigA{1.19} & $1.73$ \sigA{0.86} & $2.88$ \sigA{1.48} & $1.85$ \sigA{0.96} & $1.17$ \sigA{0.65}  &  \abref{2.16} \\
    $7.$ & \CIRCLE & \CIRCLE & \CIRCLE & \CIRCLE & $2.67$ \sigA{0.97} & $2.38$ \sigA{1.80} & $2.16$ \sigA{1.16} & $1.70$ \sigA{0.89} & $2.75$ \sigA{1.61} & $1.71$ \sigA{1.10} & $1.07$ \sigA{0.75}  & \abref{2.06} \\
    
    \midrule
    \multicolumn{13}{l}{\textit{QE-based}} \\

    $8.$ & \Circle & \Circle & \Circle & \Circle & $2.64$ & $4.87$ & $4.16$ & $1.73$ & $5.55$ & $5.39$ & $3.96$  & \abref{4.04} \\
    $9.$ & \Circle & \CIRCLE & \Circle & \Circle & $2.71$ \bad{0.07} & $4.78$ \nosig{0.09} & $4.05$ \nosig{0.11} & $1.65$ \nosig{0.07} & $5.22$ \nosig{0.33} & $5.14$ \nosig{0.25} & $4.03$ \bad{0.08}  &  \abref{3.94} \\
    $10.$ & \Circle & \Circle & \CIRCLE & \Circle & $2.11$ \sigA{0.52} & $4.33$ \sigC{0.54} & $2.82$ \sigA{1.34} & $1.30$ \sigA{0.43} & $4.49$ \sigA{1.06} & $4.31$ \sigA{1.08} & $2.89$ \sigA{1.07}  &  \abref{3.18} \\
    $11.$ & \Circle & \CIRCLE & \CIRCLE & \Circle & $2.04$ \sigA{0.59} & $4.12$ \sigC{0.75} & $3.31$ \sigB{0.85} & $1.19$ \sigA{0.54} & $4.30$ \sigA{1.25} & $4.40$ \sigA{0.99} & $3.36$ \sigC{0.60}  &  \abref{3.25} \\
    $12.$ & \CIRCLE & \CIRCLE & \Circle & \Circle & $2.26$ \sigA{0.38} & $4.18$ \sigC{0.69} & $3.50$ \sigC{0.66} & $1.54$ \nosig{0.19} & $4.60$ \sigA{0.95} & $4.62$ \sigC{0.77} & $3.73$ \nosig{0.23}  & \abref{3.49} \\
    $13.$ & \CIRCLE & \CIRCLE & \CIRCLE & \Circle & $1.90$ \sigA{0.73} & $3.39$ \sigA{1.48} & $2.76$ \sigA{1.40} & $1.23$ \sigA{0.49} & $4.17$ \sigA{1.38} & $4.12$ \sigA{1.28} & $2.97$ \sigA{0.99}  & \abref{2.93} \\
    $14.$ & \CIRCLE & \CIRCLE & \CIRCLE & \CIRCLE & $1.82$ \sigA{0.81} & $3.43$ \sigA{1.44} & $3.11$ \sigB{1.05} & $1.25$ \sigA{0.48} & $4.01$ \sigA{1.54} & $3.56$ \sigA{1.83} & $2.63$ \sigA{1.33}  &  \abref{2.83} \\

    \end{tabular}}\caption{MetricX-$23$ evaluation results of translate step-by-step and its ablation variants on the \textsc{wmt} $2023$ development datasets. We report both the reference-based and \textsc{qe}-based metric variants. Filled dots indicate active steps in the pipeline, while unfilled dots represent ablated steps. 
    When all steps are ablated, the system defaults to zero-shot translation. Colored boxes highlight performance differences compared to zero-shot: blue shades indicate {\color{blue}{significant improvements at $p<0.001$}}, green shades indicate {\color{darkseagreen}{significant improvements at $p< 0.05$}}, yellow shades indicate {\color{orange}{non-significant improvements ($p\geq0.05$)}}, while red shades indicate {\color{google-red}{non-significant regressions ($p\geq0.05$)}} against zero-shot.
    \textit{Translate step-by-step surpasses zero-shot across the board, with each step incrementally improving translation quality.}
    }\label{tab:wmt23_ablations}

\end{table*}

\paragraph{Evaluation Sets} We use \textsc{wmt} $2023$ as our \textit{\textbf{development}} set. Any prompt development and stage ablation experiments are conducted on this dataset. 
For our final \textit{\textbf{test}} set, we use the \textsc{wmt}~$2024$ datasets. 
Each of these datasets was built by translating a set of English documents into multiple languages.

Both datasets are segmented for sentence- or paragraph-level evaluation, but our approach focuses on translating with as much context as possible. Therefore, we use meta-data to merge the original segments into larger ones.  Ideally, this would result in complete documents, but current neural metrics have token-count limits beyond which they truncate their inputs. To accommodate neural evaluation, we set a maximum length of $250$ (English white-space separated) tokens each.\footnote{We also present results for a shorter set of documents, with a maximum length of $150$ tokens in Appendix~\ref{sec:shorter_documents}.}
The resulting datasets consist of $192$ documents of average token length $178$ for \textsc{wmt}~$2023$ and, $243$ documents of average token length $130$ for \textsc{wmt}~$2024$, respectively. For \textsc{wmt} $2024$ we also report per-domain results. Per-domain document counts and average lengths, as measured in English white-space separated tokens, are presented in Table~\ref{tab:per_domain_wmt24_stats}.

\paragraph{Evaluation Metrics} We evaluate our approach using MetricX-XXL-$23$~\cite{juraska-etal-2023-metricx}, the metric adopted in the most recent \textsc{wmt} $2024$ automatic evaluations. We report results on both the reference-based and the \textsc{qe}-based metric variants.
Despite being trained at the sentence level, \citet{deutsch-etal-2023-training} show that MetricX can effectively evaluate multi-sentence sequences, capped at its maximum window length.
We note that MetricX is powered by mT$5$~\cite{xue-etal-2021-mt5}, which minimizes the potential bias in favor of Gemini-generated translations.\footnote{We also report ChrF~\cite{popovic-2015-chrf} in Appendix~\ref{sec:chrf}.}
We employ paired permutation tests to determine if the observed improvements across system pairs are statistically significant.\footnote{\url{https://docs.scipy.org/doc/scipy/reference/generated/scipy.stats.permutation_test.html}}

\section{Quantitative Results}\label{sec:automatic_results}

We start by analyzing the importance of each step in the translate step-by-step pipeline. Ablation results on the \textsc{wmt} $2023$ development sets are presented in \S\ref{sec:ablation_results}. Next, the generalizability of our final step-by-step recipe is evaluated on the \textsc{wmt} $2024$ test sets in \S\ref{sec:wmt24_results}, with comparison to prior work in \S\ref{sec:prior_work}.

\subsection{Analyzing Step Importance}\label{sec:ablation_results}

Automatic evaluation results on our development sets are presented in Table~\ref{tab:wmt23_ablations}. Overall, translation artifacts extracted through the step-by-step process yield consistently better document translations compared to the zero-shot mode according to both reference-  (lines $3$--$7$ vs.\ $1$) and \textsc{qe}-based (lines $10$--$14$ vs.\ $8$) versions of MetricX. Ablating the various steps from the pipeline gives insights into how each step contributes to the overall quality improvements. We describe those below.

\paragraph{Importance of Pre-translation Research}
Modelling pre-translation processes is crucial for achieving higher quality translations compared to the zero-shot. Simply prompting for a draft translation without asking for pre-translation research yields only small and non-significant improvements or even regressions over the zero-shot (lines $2$ vs.\ $1$ and $9$ vs.\ $8$). This result rules out the possibility that any observed improvements are solely due to a better prompt for the draft translation, which was modified to emphasize faithfulness to the source (\S\ref{sec:method_draft}). However, combining the research and draft steps achieves consistently higher quality translations compared to zero-shot (lines $5$ vs.\ $1$ and $12$ vs.\ $8$). Importantly, those improvements are consistently statistical significant ($p < 0.0001$) across languages (measured by reference-based metrics), except for Hebrew, which shows non-significant improvements compared to zero-shot ($p \geq 0.05$).

\paragraph{Importance of Refinement}

Moving to the evaluation of the \textit{refined} document translations, we notice an interesting trend. The refinement step consistently improves the translation quality, regardless of the initial translation it processes, i.e., the zero-shot (lines $3$ vs.\ $1$ and $10$ vs.\ $8$), the single-turn draft (lines $4$ vs.\ $2$ and $11$ vs.\ $9$), and the research-informed draft (lines $6$ vs.\ $5$ and $13$ vs.\ $12$). This demonstrates that the effectiveness of the refinement stage is not conditioned on the initial translation. However, the strongest quality improvements---reaching consistently high levels of statistical significance ($p <0.001$) over the zero-shot translations---are observed when the refinement stage is combined with
the pre-translation research (lines $6$ vs.\ $1$ and $13$ vs.\ $8$), highlighting that those stages bring complimentary benefits.

\paragraph{Importance of Proofreading}
Finally, the evaluation of the \textit{proofreading} document translations, 
indicate that this stage contributes modest average improvements (lines $7$ vs.\ $6$ and $14$ vs.\ $13$). Unlike previous stages, the impact of proofreading appears to be more language dependent. Ukrainian stands out as the only language that clearly benefits from a proofreading stage, while others show only minor differences in quality compared to their refined translations. 

\subsection{Generalizability of Step-by-Step}\label{sec:wmt24_results}
\begin{table*}[!ht]
    \centering
    \scalebox{0.59}{
    \begin{tabular}{llllllllll>{\columncolor[gray]{0.9}}c}
    \arrayrulecolor{gray!40}
    \rowcolor{blue!10}
    &  \textbf{\textsc{de}} & \textbf{\textsc{es}} & \textbf{\textsc{zh}} & \textbf{\textsc{ru}} & \textbf{\textsc{uk}} & \textbf{\textsc{ja}} & \textbf{\textsc{hi}} & \textbf{\textsc{is}} & \textbf{\textsc{cs}} & \textbf{\textsc{average}}\\
    
    \multicolumn{11}{r}{\textit{Ref-based}} \\

    \textit{Zero-shot} &  $1.90$ & $3.23$ & $3.48$ & $3.02$ & $3.15$ & $2.29$ & $3.65$ & $4.01$ & $2.65$  & \xxxs{3.04} \\
    \textsc{sbys}: \textit{Research \& Drafting} &  $1.68$ \sigA{0.22} & $2.69$ \sigA{0.54} & $2.99$ \sigA{0.49} & $2.53$ \sigA{0.49} & $2.81$ \sigA{0.35} & $1.92$ \sigA{0.37} & $2.52$ \sigA{1.13} & $3.77$ \sigB{0.24} & $2.30$ \sigC{0.35}  &  \xxxs{2.58} \\
    \textsc{sbys}: \textit{Refinement}           &  $1.45$ \sigA{0.45} & $2.29$ \sigA{0.94} & $2.45$ \sigA{1.03} & $2.21$ \sigA{0.81} & $2.58$ \sigA{0.57} & $1.64$ \sigA{0.66} & $2.31$ \sigA{1.35} & $3.14$ \sigA{0.87} & $2.10$ \sigB{0.55}  &  \xxxs{2.24} \\
    \textsc{sbys}: \textit{Proofreading}         & $1.35$  \sigA{0.54} & $2.27$ \sigA{0.96} & $2.42$ \sigA{1.06} & $2.21$ \sigA{0.81} & $2.49$ \sigA{0.66} & $1.67$ \sigA{0.62} & $2.09$ \sigA{1.56} & $3.15$ \sigA{0.86} & $2.14$ \sigC{0.51}  &  \xxxs{2.20} \\
    
    \vspace{0.01cm}\\
    \hline
    \multicolumn{11}{r}{\textit{QE-based}} \\
        
    \textit{Zero-shot} & $1.97$ & $2.59$ & $2.23$ & $1.87$ & $2.23$ & $1.32$ & $4.81$ & $3.47$ & $2.08$  &  \xxxs{2.51} \\   

    \textsc{sbys}: \textit{Research \& Drafting} & $1.72$ \sigA{0.25} & $2.23$ \sigA{0.36} & $2.08$ \nosig{0.15} & $1.54$ \sigA{0.33} & $1.81$ \sigA{0.41} & $1.19$ \sigB{0.13} & $4.12$ \sigB{0.69} & $3.43$ \nosig{0.04} & $1.97$ \nosig{0.11}  &  \xxxs{2.23} \\
     \textsc{sbys}: \textit{Refinement}          & $1.38$ \sigA{0.59} & $1.78$ \sigA{0.81} & $1.71$ \sigA{0.52} & $1.21$ \sigA{0.66} & $1.34$ \sigA{0.89} & $0.95$ \sigA{0.37} & $3.47$ \sigA{1.34} & $2.79$ \sigA{0.68} & $1.51$ \sigA{0.56}  & \xxxs{1.79} \\
    \textsc{sbys}: \textit{Proofreading}         & $1.25$ \sigA{0.72} & $1.74$ \sigA{0.84} & $1.63$ \sigA{0.60} & $1.14$ \sigA{0.73} & $1.32$ \sigA{0.91} & $0.93$ \sigA{0.40} & $3.35$ \sigA{1.46} & $2.65$ \sigA{0.82} & $1.45$ \sigA{0.63}  &  \xxxs{1.72} \\

    \end{tabular}}\caption{MetricX-$23$ results comparing step-by-step (\textsc{sbys}) with zero-shot on the \textsc{wmt} $2024$ test datasets. When all steps are ablated, the system defaults to zero-shot translation.  Colored boxes highlight performance differences compared to zero-shot: blue shades indicate {\color{blue}{significant improvements at $p<0.001$}}, green shades indicate {\color{darkseagreen}{significant improvements at $p< 0.05$}}, while yellow shades indicate {\color{orange}{non-significant improvements~($p\geq0.05$)}}. \textit{Translate step-by-step surpasses zero-shot, with each step incrementally improving translation quality.}}\label{tab:wmt24_automatic}

\end{table*}

\begin{figure}[!ht]
    \centering
    \includegraphics[scale=0.445]{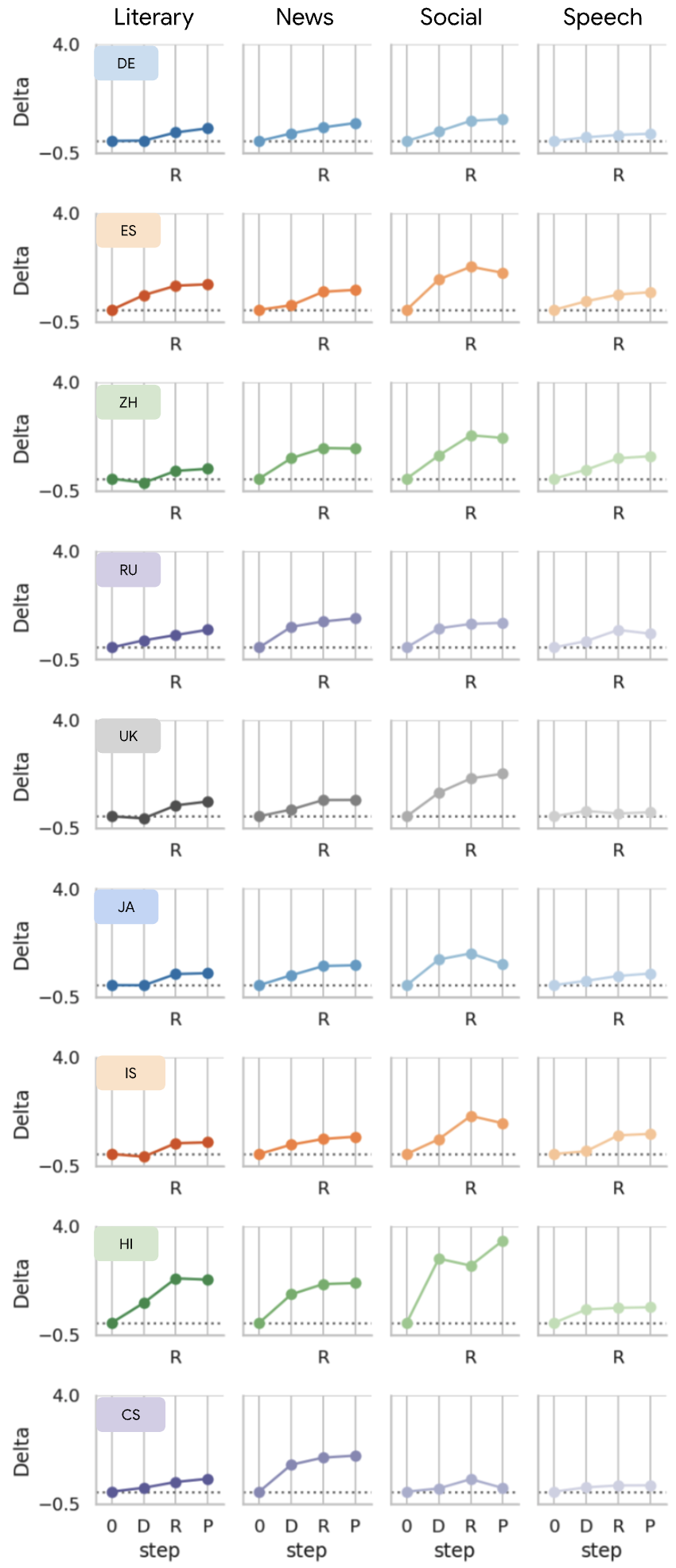}
    \caption{Domain-level comparison between zero-shot and step-by-step translations on \textsc{wmt} $2024$ using reference-based MetricX-$23$. Each data point represents the delta from zero-shot (dotted horizontal line). The steps are denoted as follows: 0 (zero-shot), D (draft after research), R (refinement), and P (proofreading).}
    \label{fig:per_domain_metrics}
\end{figure}

Table~\ref{tab:wmt24_automatic} presents results on the \textsc{wmt} $2024$ test set. Across the board, translating step-by-step exhibits the same trends noticed on our development set (as discussed in \S\ref{sec:ablation_results}). This confirms the generalizability of our proposed approach, crucially, on a wider range of languages. Concretely, the draft translations outperform the zero-shot translations. The refined stages bring additional quality improvements across the board, with the proofreading stage contributing small improvements for most languages.

To better understand the robustness of our approach we present a per-domain analysis in
Figure~\ref{fig:per_domain_metrics}. As shown, translation quality improvements of step-by-step translations over zero-shot are observed across all domains, with speech showing the least and social the most significant gains.

\subsection{Contextualizing Step-by-Step Gains}\label{sec:prior_work} 

Having demonstrated how translate step-by-step improves long-form translation with \textsc{llm}s over zero-shot translation, we now contextualize these gains by comparing our approach to two representative baselines: a) methods that leverage non-parametric knowledge for best translation selection, and b) segment-level baselines that translate documents using the pre-defined segmentation provided in \textsc{wmt} $2024$ test sets.  

\paragraph{Conditions} As a representative of the first class, we compare against \textsc{maps}~\cite{He2023ExploringHT}. This baseline employs an \textsc{llm} to analyze the source text for topic, keywords, and similar examples, generating three candidate translations conditioned on each knowledge type. Then, a \textsc{qe} metric selects the best candidate. To create a fair comparison, we re-implement their method using Gemini $1.5$ Pro, using the prompts provided in their released code. To create an even stronger baseline, we perform candidate selection with the \textsc{qe} variant of MetricX-$23$, which we know correlates well with the final reference-based MetricX-$23$, creating an advantage for \textsc{maps}.

For the second class of baselines, we consider two approaches: a) zero-shot translation applied to each segment individually using Gemini $1.5$ Pro, both with (\textsc{zero-shot in context}) and without (\textsc{zero-shot}) access to the full document in the input prompt, and b) a comparison with the segment-level translations from Unbabel-Tower$70$B, the top-performing system of \textsc{wmt} $2024$ based on early automatic evaluations~\cite{Kocmi2024PreliminaryWR}. To get comparable document-level metrics, before evaluation, we concatenate the segment-level translation back into the mini-documents, as described in \S\ref{sec:doc_setting}. 

We focus our comparisons on \textsc{en-de}, \textsc{en-ja}, \textsc{en-zh}, as \textsc{maps} requires in-context demonstrations that were made available only for those languages by the original authors. For a fair comparison with Unbabel-Tower$70$B, we exclude the speech domain from our comparison, as \textsc{wmt} $2024$ submissions were given \textsc{asr} transcripts instead of human-sourced transcripts in this domain.

\begin{table}[!t]
    \centering
    \scalebox{0.59}{
    \begin{tabular}{lcccc}
    \rowcolor{gray!20}
    \textbf{\textsc{method}} & \textsc{\textbf{doc.}}    & \textbf{\textsc{en-de}} & \textbf{\textsc{en-zh}} & \textbf{\textsc{en-ja}}  \\
    \textsc{unbabel-tower70b}       & \xmark   & \clustersigde{1} \priorde{1.42} &  \clustersigzh{1}  \priorja{2.77} & \clustersig{2} \priorzh{2.16} \\
    \textsc{zero-shot}              & \xmark   & \clustersigde{2}  \priorde{1.98} & \clustersigzh{3} \priorja{3.65} & \clustersig{3} \priorzh{2.60} \\
    \textsc{zero-shot in context}   & \xmark & \clustersigde{2} \priorde{1.86} & \clustersigzh{2} \priorja{3.33} & \clustersig{2} \priorzh{2.19} \\
    \textsc{zero-shot}              & \bluecheck      & \clustersigde{3}  \priorde{2.02} & \clustersigzh{3} \priorja{3.91} & \clustersig{3} \priorzh{2.47} \\
    \textsc{maps}                    & \bluecheck & \clustersigde{2} \priorde{1.91} & \clustersigzh{2} \priorja{3.25} & \clustersig{2} \priorzh{2.19} \\
     \textsc{sbys}: \textit{Research \& Drafting} & \bluecheck &  \clustersigde{2}  \priorde{1.75} & \clustersigzh{2} \priorja{3.32} & \clustersig{2} \priorzh{1.94} \\
      \textsc{sbys}: \textit{Refinement} & \bluecheck & \clustersigde{1}  \priorde{1.41}	& \clustersigzh{1}  \priorja{2.73} & 	\clustersig{1} \priorzh{1.58} \\
     \textsc{sbys}: \textit{Proofreading} & \bluecheck & \clustersigde{1} \priorde{1.27} & \clustersigzh{1} \priorja{2.75} & \clustersig{1} \priorzh{1.73} \\
    \end{tabular}}
    \caption{Comparison of step-by-step (\textsc{sbys}) with representative baselines (lower scores are better) on \textsc{wmt} $2024$ according to Metric-X (reference-based). The second column indicates whether translation is performed on the entire document or by merging segment-level translations. Numbered squares represent significance clusters~\cite{freitag-etal-2023-results} at $p=0.05$. \textit{Translate step-by-step matches or exceeds all compared baselines, crucially, without accessing external resources.}}
    \label{tab:prior_work_table}
\end{table}

\paragraph{Results} Table~\ref{tab:prior_work_table} compares step-by-step against various baselines.
Notably, even the initial stage, where the draft translation is conditioned on cross-lingual research 
(\textsc{sbys}: \textit{Research \& Drafting}) demonstrates competitive performance against \textsc{maps}, falling within the same statistical significance cluster. This highlights the effectiveness of our pre-translation strategy compared to the background information used by \textsc{maps}.
Comparing the final, proofreading stage of step-by-step (\textsc{sbys}: \textit{Proofreading}) with \textsc{maps} reveals significant translation quality gains: $0.64$ improvement for \textsc{de}, $0.50$ for \textsc{zh}, and $0.46$ for \textsc{ja}. 
Notably, these improvements are achieved even though \textsc{maps} uses the same \textsc{qe} model family as MetricX for final candidate selection, giving it an inherent advantage. In contrast, \textsc{sbys} relies solely on the model's internal, parametric knowledge throughout the entire translation process.

Comparing the final, proofreading stage of step-by-step with the segment-level baselines helps put the improvements in perspective. Concretely, the segment-level zero-shot baselines (second and third lines in Table~\ref{tab:prior_work_table}) fall significantly behind the step-by-step final translations (\textsc{sbys}:~\textit{Proofreading})
across all languages by more than $0.7$ and $0.4$ MetricX points when compared to the out-of- and in-context variants, respectively. This demonstrates that simply translating documents at a finer granularity is not sufficient for boosting the \textsc{llm}'s translation quality. 

Finally, comparing the final, proofreading stage of our approach with the merged translations from Unbabel-Tower70B, reveals that our approach achieves statistically comparable performance for Chinese and German  ($0.02$ and $0.15$ improvements respectively) and significantly better performance for 
 Japanese ($0.43$ improvement). These improvements over the top-performing \textsc{wmt} $2024$ system demonstrate the competitiveness of the step-by-step approach, especially given that the competing system relies on external \textsc{qe} metrics and computationally expensive decoding strategies to improve translation quality.

\section{Qualitative Analysis}\label{sec:qualitative_analysis}

We conduct a qualitative analysis on a small subset of model outputs from all stages to understand the strengths and weaknesses of our step-by-step approach. To this end, we first compute the score difference between the final translation and the zero-shot output on \textsc{wmt} $2024$ English to Chinese, and then randomly sample up to 5 examples from either end (i.e., examples for which the final translation quality either substantially improves or degrades over the zero-shot baseline).\footnote{The exact sample ranges of the score difference are [-6, -2] and [1, 6]. Examples from beyond these ranges typically demonstrate clear signs of model degeneration and are therefore excluded from this analysis.} One of the authors (native speaker of Chinese) manually inspected the sampled outputs and took notes on the salient properties of the pre-translation artifacts and the incremental changes from the different stages of the step-by-step process.

\begin{table*}[!ht]
    \centering
    \scalebox{0.7}{
    \begin{tabular}{p{0.05\linewidth}p{0.5\linewidth}p{0.55\linewidth}p{0.2\linewidth}}
    \arrayrulecolor{gray!40}
    \toprule
\rowcolor{gray!10}
& \textsc{\textbf{source snippet}} & \textsc{\textbf{pre-translation research}} & \textbf{\textsc{notes}}\\
\bottomrule
\rowcolor{blue!3}
\vspace{0.01cm} $1$. \bluecheck & \vspace{0.01cm} trying my hand at miniatures \newline redid the head! \newline \contentOmitted \newline @user26 cheeked up. I love HER* 
 & \vspace{0.01cm} \textbf{``cheeked up''}: This slang term likely refers to making the miniature's cheeks fuller or more prominent. A direct translation wouldn't be accurate.
& \vspace{0.01cm} \textit{\textcolor{blue}{Understanding source correctly in the context of making miniatures.}}\\
\bottomrule
\rowcolor{blue!3}
\vspace{0.01cm} $2$. \bluecheck & \vspace{0.01cm}
\#TeaganAir \#pilot \#flying \#pa28 \#PilotTraining \newline \contentOmitted \newline Last week the weather didn't cooperate so we're back at it today.
& \vspace{0.01cm}
\textbf{``The weather didn't cooperate''}: This is a figurative way of saying the weather was not suitable for flying. & \vspace{0.01cm} \textit{\textcolor{blue}{Good example of indirect translation.}} \\
\bottomrule
\rowcolor{blue!3}
\vspace{0.01cm} $3$. \bluecheck & \vspace{0.01cm}
It is nice to have art and entertainment to help us make sense of our life. \newline @user32 I could perform a liver transplant thanks to Hospital Playlist.
& \vspace{0.01cm}
\textbf{``I could perform a liver transplant thanks to Hospital Playlist''}: This is a humorous exaggeration, implying that the show provided a lot of information about liver transplants. & \vspace{0.01cm}
\textit{\textcolor{blue}{Good humor detection.}} \\
\bottomrule
\rowcolor{alizarin!4}
\vspace{0.01cm} $4$. \xmark & \vspace{0.01cm}
He was covered in cuts and scratches and seemed a bit dazed, but he didn't seem too badly injured.
&\vspace{0.01cm}
\textbf{``a bit dazed''}: This describes a state of confusion or disorientation. A literal translation might not convey the same meaning. & \vspace{0.01cm}
\textit{\textcolor{alizarin}{Literal translation would do just fine.}}\\
\bottomrule
\rowcolor{alizarin!4}
\vspace{0.01cm} $5$. \xmark & 
\vspace{0.01cm} been really enjoying kneadatite (green stuff), it's like sculpting with bubble gum but you get used to that and it's tough and flexible when it cures
&
\vspace{0.01cm} \textbf{``like sculpting with bubble gum''}: : This is a simile comparing the texture of Kneadatite to bubble gum. It needs to be translated in a way that makes sense to a Chinese audience who might not be familiar with the specific texture of bubble gum. & \vspace{0.01cm}
\textit{\textcolor{alizarin}{Strange commentary on contextualizing the texture of bubble gum to a Chinese audience.}}\\
\bottomrule
    \end{tabular}}
    \caption{Samples of pre-translation research outputs along with notes from the author-annotator.}
    \label{tab:example_outputs_main}
\end{table*}


\paragraph{Pre-drafting} For pre-drafting research, we observe that the \textsc{llm} is highly capable of understanding the source in a wide variety of contexts. As showcased in Table~\ref{tab:example_outputs_main}, the \textsc{llm}  is able to correctly interpret slang (example 1: \textit{cheeked up} in the context of making miniatures), recognize figurative usage (example 2: \textit{the weather didn't cooperate} in the context of flying a plane), and detect humorous expressions (example 3). This strength is especially pronounced when even the references show clear signs of human translators misinterpreting the source (see the next subsection for full examples).

On the other hand, the \textsc{llm}  is also prone to over-generate and seems too eager to confirm with the given instruction to find instances of indirect translation. This resulted in false positives where a direct and literal translation is already adequate (example 4: \textit{a bit dazed} can be directly translated into Chinese), and in some cases bizarre cultural commentaries (example 5 for asking to contextualize the texture of bubble gum).

\paragraph{Translations} The observed understanding of the source texts seems to directly contribute to more fluent and context-appropriate translations. Table~\ref{tab:example_outputs_comparison} in \S\ref{sec:more_example_outputs} shows several interesting examples. There are quite a few instances where the step-by-step approach produces the correct translation even when the reference fails to recognize the context the phrase appears in. For example, our method correctly interprets the meaning of \textit{cheeked up} in the first example to be ``having a full cheek'' when the zero-shot translates it to \textit{blushing} and the reference translates it to \textit{talking nonsense}. Similarly for the second example, the term \textit{threading} is correctly understood as a thread of posts on a social media platform by all step-by-step translations, whereas the zero-shot interpretes it as a computing terminology (as in \textit{multithreading}) and the reference interprets it as \textit{study}. 

It is also evident that the refinement improves the fluency significantly. The third example shows that both the zero-shot and the draft translates the source literally. This entails preserving the original source structure and translating the source word \textit{ridiculous} directly. The result is somewhat awkward and sounds like translationese, which is particularly jarring considering the social media domain of the source text. This issue is rectified by both the refined translation and the reference. On the other hand, the refinement process is not perfect and does regress to less fluent outputs at times (example 4). Another prominent failure mode typical of all \textsc{llm} outputs is hallucinations. In the fifth example, the refinement adds ``touching the bruise'', which is not present in the source.

\section{Conclusion}

We introduce a step-by-step approach to long-form text translation using \textsc{llm}s. Inspired by literature on translation studies, we decompose the translation process into distinct stages, modeling pre-translation research, drafting, refinement, and proofreading though a multi-turn interaction with Gemini $1.5$ Pro. 
Extensive automatic evaluations on \textsc{wmt} $2023$ and \textsc{wmt} $2024$ tasks in ten languages demonstrate that our approach improves  translation quality over directly translating the entire document with a single prompt.

Furthermore, comparison with competitive baselines, including similar human-like \textsc{llm}-driven approaches and top-performing systems that employ segment-by-segment translation of a document, reveals the strong performance of our approach. Our findings highlight the potential of \textsc{llm}s to progressively improve their translations, 
moving beyond the traditional view of machine translation as a monolithic sequence mapping task. 

\section*{Limitations}

While our study reveals promising step-by-step improvements across various languages and domains, we acknowledge the limitations of solely relying on automatic metrics for evaluation. While metric improvements give us a  consistent signal, human evaluation is needed to further validate the effectiveness of the approach and reveal a more nuanced understanding of the translation properties introduced at each step. 
We also acknowledge that our analysis is based solely on one family of metrics, due to context window limitations of other neural metrics in evaluating longer texts. 

Finally, our pipeline is developed and tested solely on Gemini. Since different \textsc{llm}s might exhibit different instruction-following capabilities across languages, the generalizability of this approach to other \textsc{llm}s requires further investigation.

\section*{Ethics Statement}

This paper explores the use of \textsc{llm}s to improve translation quality. In doing so, our approach starts from an initial translation that prioritizes faithfulness to the source text.
Subsequent stages focus on improving fluency which, as they deviate more from the source, increase the risk of hallucinations~\cite{guerreiro-etal-2023-hallucinations}---a critical issue in machine translation, potentially leading to misleading translations.

Moreover, the increasing fluency of machine translations presents new challenges  when prioritized over adequacy~\cite{martindale-carpuat-2018-fluency}, as users might trust their outputs blindly, even when incorrect. This highlights the need for careful adoption of those translation systems and the developing of strategies that help users calibrate their trust appropriately.

\bibliography{anthology,custom}
\bibliographystyle{acl_natbib}

\appendix
\newpage
\section{Appendices}

\begin{table*}[!ht]
    \centering
    \scalebox{0.67}{
    \begin{tabular}{rlllllllllll>{\columncolor[gray]{0.9}}c}
    \arrayrulecolor{gray!20}
    \rowcolor{blue!10}
     & \rot{\textit{Research}} & \rot{\textit{Draft}} & \rot{\textit{Refinement}} & \rot{\textit{Proofreading}} & \textbf{\textsc{zh}} & \textbf{\textsc{uk}} & \textbf{\textsc{ru}} & \textbf{\textsc{ja}} & \textbf{\textsc{he}} & \textbf{\textsc{cs}} & \textbf{\textsc{de}} & \textbf{\textsc{average}} \\

$1.$ & \Circle & \Circle & \Circle & \Circle & $48.04$ & $61.85$ & $63.55$ & $38.75$ & $64.03$ & $67.62$ & $71.81$  &  $59.38$ \\
$2.$ & \Circle & \CIRCLE & \Circle & \Circle & $48.69$ \goodchrfS{0.65} & $61.81$ \badchrfS{0.04} & $63.93$ \goodchrfS{0.38} & $39.00$ \goodchrfS{0.25} & $64.68$ \goodchrfS{0.65} & $67.63$ \goodchrfS{0.01} & $71.79$ \badchrfS{0.02}  &  $59.65$ \\
$3.$ & \Circle & \Circle & \CIRCLE & \Circle & $41.48$ \badchrf{6.56} & $59.44$ \badchrf{2.41} & $59.33$ \badchrf{4.22} & $36.19$ \badchrf{2.56} & $60.26$ \badchrf{3.77} & $63.44$ \badchrf{4.18} & $66.89$ \badchrf{4.92}  &  $55.29$ \\
$4.$ & \Circle & \CIRCLE & \CIRCLE & \Circle & $43.14$ \badchrf{4.90} & $59.58$ \badchrf{2.27} & $60.37$ \badchrf{3.18} & $37.45$ \badchrfS{1.30} & $60.92$ \badchrf{3.11} & $63.04$ \badchrf{4.58} & $68.71$ \badchrf{3.10}  &  $56.17$ \\
$5.$ & \CIRCLE & \CIRCLE & \Circle & \Circle & $45.98$ \badchrf{2.06} & $61.51$ \badchrfS{0.34} & $63.04$ \badchrfS{0.51} & $39.30$ \goodchrfS{0.55} & $62.89$ \badchrfS{1.14} & $67.17$ \badchrfS{0.45} & $71.07$ \badchrfS{0.74}  &  $58.71$ \\
$6.$ & \CIRCLE & \CIRCLE & \CIRCLE & \Circle & $41.03$ \badchrf{7.01} & $58.72$ \badchrf{3.13} & $59.44$ \badchrf{4.11} & $37.65$ \badchrfS{1.10} & $59.91$ \badchrf{4.12} & $63.02$ \badchrf{4.60} & $67.61$ \badchrf{4.20}  &  $55.34$ \\
$7.$ & \CIRCLE & \CIRCLE & \CIRCLE & \CIRCLE & $40.71$ \badchrf{7.33} & $58.78$ \badchrf{3.07} & $59.23$ \badchrf{4.32} & $37.51$ \badchrfS{1.24} & $59.65$ \badchrf{4.38} & $63.11$ \badchrf{4.51} & $67.49$ \badchrf{4.32}  &  $55.21$ \\

    \end{tabular}}\caption{ChrF evaluation results of translate step-by-step and its ablation variants on the \textsc{wmt} $2023$ development datasets. Filled dots indicate active steps in the pipeline, while unfilled dots represent ablated steps. 
    When all steps are ablated, the system defaults to zero-shot translation
    }\label{tab:wmt23_ablations_chrf}

\end{table*}

\begin{table*}[!ht]
    \centering
    \scalebox{0.55}{
    \begin{tabular}{llllllllll>{\columncolor[gray]{0.9}}c}
    \arrayrulecolor{gray!40}
    \rowcolor{blue!10}
    &  \textbf{\textsc{de}} & \textbf{\textsc{es}} & \textbf{\textsc{zh}} & \textbf{\textsc{ru}} & \textbf{\textsc{uk}} & \textbf{\textsc{ja}} & \textbf{\textsc{hi}} & \textbf{\textsc{is}} & \textbf{\textsc{cs}} & \textbf{\textsc{average}}\\
    
 \textit{Zero-shot} &  $65.48$ & $72.96$ & $44.21$ & $55.51$ & $59.90$ & $39.75$ & $55.94$ & $53.23$ & $60.81$  &  $56.42$ \\
 \textit{Research \& Drafting} &  $64.67$ \badchrfS{0.81} & $72.30$ \badchrfS{0.66} & $42.73$ \badchrfS{1.48} & $57.30$ \goodchrfS{1.79} & $60.06$ \goodchrfS{0.16} & $41.19$ \goodchrfS{1.44} & $56.16$ \goodchrfS{0.22} & $53.09$ \badchrfS{0.14} & $60.31$ \badchrfS{0.50}  &  $56.42$ \\
 \textit{Refinement} &  $61.72$ \badchrf{3.76} & $69.22$ \badchrf{3.74} & $38.26$ \badchrf{5.95} & $55.09$ \badchrfS{0.42} & $57.25$ \badchrf{2.65} & $39.15$ \badchrfS{0.60} & $52.60$ \badchrf{3.34} & $52.62$ \badchrfS{0.61} & $57.29$ \badchrf{3.52}  &  $53.69$ \\
 \textit{Proofreading} &  $61.62$ \badchrf{3.86} & $69.04$ \badchrf{3.92} & $38.41$ \badchrf{5.80} & $54.96$ \badchrfS{0.55} & $57.14$ \badchrf{2.76} & $38.87$ \badchrfS{0.88} & $53.47$ \badchrf{2.47} & $52.32$ \badchrfS{0.91} & $56.98$ \badchrf{3.83}  &  $53.65$ \\
    \end{tabular}}\caption{ChrF results comparing step-by-step with zero-shot performance on the \textsc{wmt} $2024$ test datasets.}\label{tab:wmt24_automatic_chrf}

\end{table*}

\begin{table*}[!ht]
    \centering
    \scalebox{0.6}{
    \begin{tabular}{llllllllll>{\columncolor[gray]{0.9}}c}
    \arrayrulecolor{gray!40}
    \rowcolor{blue!10}
    &  \textbf{\textsc{de}} & \textbf{\textsc{es}} & \textbf{\textsc{zh}} & \textbf{\textsc{ru}} & \textbf{\textsc{uk}} & \textbf{\textsc{ja}} & \textbf{\textsc{hi}} & \textbf{\textsc{is}} & \textbf{\textsc{cs}} & \textbf{\textsc{average}}\\
    
    \multicolumn{11}{r}{\textit{Ref-based}} \\

    \textit{Zero-shot} &  $1.89$ & $3.10$ & $3.25$ & $2.90$ & $2.99$ & $2.31$ & $3.03$ & $3.79$ & $2.37$  &  $2.85$ \\
    \textit{Research \& Drafting} &  $1.67$ \goodS{0.23} & $2.61$ \goodM{0.49} & $2.80$ \goodM{0.45} & $2.53$ \goodM{0.37} & $2.67$ \goodM{0.32} & $1.91$ \goodM{0.40} & $2.00$ \goodXL{1.03} & $3.45$ \goodM{0.34} & $2.16$ \goodS{0.21}  &  $2.42$ \\
    \textit{Refinement} &  $1.44$ \goodM{0.45} & $2.20$ \goodL{0.90} & $2.33$ \goodL{0.92} & $2.17$ \goodL{0.73} & $2.34$ \goodL{0.65} & $1.61$ \goodL{0.70} & $1.62$ \goodXL{1.41} & $3.02$ \goodL{0.76} & $2.00$ \goodM{0.37}  &  $2.08$ \\
    \textit{Proofreading} &  $1.36$ \goodL{0.53} & $2.11$ \goodL{0.99} & $2.28$ \goodL{0.97} & $2.20$ \goodL{0.69} & $2.27$ \goodL{0.72} & $1.65$ \goodL{0.66} & $1.60$ \goodXL{1.43} & $3.04$ \goodL{0.75} & $1.98$ \goodM{0.39}  &  $2.05$ \\

    \vspace{0.01cm}\\
    \hline
    \multicolumn{11}{r}{\textit{QE-based}} \\
    \textit{Zero-shot} &  $1.81$ & $2.34$ & $2.13$ & $1.67$ & $1.96$ & $1.26$ & $1.98$ & $3.15$ & $1.85$  &  $2.02$ \\
    \textit{Research \& Drafting} &  $1.62$ \goodS{0.18} & $2.03$ \goodM{0.31} & $1.85$ \goodS{0.28} & $1.40$ \goodS{0.26} & $1.60$ \goodM{0.36} & $1.10$ \goodS{0.15} & $1.40$ \goodL{0.58} & $2.93$ \goodS{0.21} & $1.73$ \goodS{0.12}  &  $1.74$ \\
    \textit{Refinement} &  $1.24$ \goodL{0.57} & $1.61$ \goodL{0.73} & $1.51$ \goodL{0.62} & $1.05$ \goodL{0.62} & $1.17$ \goodL{0.79} & $0.91$ \goodM{0.35} & $0.96$ \goodXL{1.01} & $2.37$ \goodL{0.78} & $1.31$ \goodL{0.53}  &  $1.35$ \\
   \textit{Proofreading} &  $1.12$ \goodL{0.68} & $1.54$ \goodL{0.80} & $1.44$ \goodL{0.69} & $0.99$ \goodL{0.67} & $1.10$ \goodL{0.86} & $0.88$ \goodM{0.38} & $0.92$ \goodXL{1.06} & $2.24$ \goodL{0.90} & $1.22$ \goodL{0.62}  &  $1.27$ \\
    \end{tabular}}\caption{MetricX-23 evaluation results comparing step-by-step with zero-shot performance on the \textsc{wmt} $2024$ test datasets, where each document has a maximum length of $150$ tokens. \textit{Translate step-by-step surpasses zero-shot, with each step incrementally improving translation quality.}}\label{tab:wmt24_automatic_150}

\end{table*}

\begin{table}[!t]
    \scalebox{0.86}{
    \centering
    \begin{tabular}{lrrrr}
    \rowcolor{gray!20}
    Domain            & Literary & News & Social & Speech \\ 
    \# Docs.          & $66$  & $73$  & $75$   & $112$\\
    Avg. Length       & $120$ & $110$ & $105$  & $72$\\
    \end{tabular}}
    \caption{Per-domain statistics for \textsc{wmt} $2024$, when blobbing with $150$ max for total of $327$ docs.}
    \label{tab:per_domain_wmt24_stats_150}
\end{table}

\subsection{Results on Shorter Documents}\label{sec:shorter_documents}

Table~\ref{tab:wmt24_automatic_150} presents automatic evaluation results of step-by-step on shorter documents, where segments are grouped together such that they do not exceed a token limit of $150$ white-space separated tokens. The dataset statistics are presented in Table~\ref{tab:per_domain_wmt24_stats_150}. We observe the same trends with the ones reported with larger documents in \S\ref{sec:automatic_results}.

\subsection{Results on ChrF}\label{sec:chrf}

Tables~\ref{tab:wmt23_ablations_chrf} and \ref{tab:wmt24_automatic_chrf} report ChrF scores on \textsc{wmt} $2023$ and \textsc{wmt} $2024$, respectively. 
As anticipated with string-based metrics, \textsc{llm} translations which prioritize fluency receive lower scores compared to those that are by construction instructed to be closer to the source text. This behavior is in line with observations of prior work that employ similar human-like translation strategies with \textsc{llm}s~\cite{Wu2024PerhapsBH}.


\subsection{Prompts}\label{sec:appendix_prompts}
Tables~\ref{tab:sbys_prompts} and \ref{tab:baseline_prompts} present the complete prompts we used for our translate step-by-step framework and baselines. It has come to our attention that the prompts used in the experiments contain a few typographical errors. Preliminary results using revised prompts show comparable, if not slightly improved results (in the range of $0.1-0.2$ MetricX-23 score points), across all steps.

\begin{table*}[!ht]
    \centering
    \scalebox{0.8}{
    \begin{tabular}{p{1.\linewidth}}
    \hline
   \rowcolor{gray!10}
    \textbf{\textsc{pre-translation research}}\\   
    You will be asked to translate a piece of text form \textcolor{alizarin}{English} into \textcolor{alizarin}{Chinese} following the five stages of the translation process. Here is the context in which the text appears: \\
     \\
     Context: \textit{\textcolor{darkseagreen}{placeholder source text}}\\
     \\
     To start, let's do some pre-drafting research on the above context: \\
     \\
     \textbf{Research:} \\
     During this phase, thorough research is essential to address components of the context text that pose translation challenges. The goal is to establish a comprehensive translation plan that covers the following category:\\
     \begin{itemize}
         \item[*] \textbf{Idiomatic Expressions:}
         \begin{itemize}
             \item[*] Identify idiomatic expressions that cannot be directly translated word-for-word into  \textcolor{alizarin}{Chinese}.
         \end{itemize}
     \end{itemize}
     \\
    \hline
   \rowcolor{gray!10}
    \textbf{\textsc{drafting}}\\ 
    
  Now, let's move on to the drafting stage.\\
  \\
\textbf{Draft Translation:}\\
In this phase, your primary objective is to create a draft translation that accurately conveys the meaning of the source text presented below. At this stage, it is crucial to focus on adequacy, ensuring that your translation closely adheres to the source text. Your response should conclude with the draft translation. If context is missing, generate a general translation that is adaptable to various contexts. Avoid adding any additional information not present in the source text. All elements of the source text should be present in the translation.
\\
\\
Give your best one translation for the following piece of text based on the pre-drafting analysis without providing alternatives:
\\
\\
\textcolor{alizarin}{English}: \textit{\textcolor{darkseagreen}{placeholder source text}}
\\
\\
    \hline
   \rowcolor{gray!10}
    \textbf{\textsc{refinement}}\\ 
Now let's move to the next stage.\\
\\
\textbf{Post-editing with local refinement}\\
In this stage, the primary aim is to refine the draft translation by making micro-level improvements that improve the draft's fluency.
\\
\\
Provide only one refined translation and do not output anything else after that.
\\
\\
    \hline
   \rowcolor{gray!10}
    \textbf{\textsc{proofreading}}\\
You are tasked with proofreading a translation that has been revised for improved fluency. The refined translation has been generated by editing the draft translation.\\
\\
\textbf{Proofreading and Final Editing}\\
The goal is to provide a polished final translation of the source text. For you reference, below are the source text, the draft, and refined translations.\\
\\
\textbf{Source Text}\\
\textit{\textcolor{darkseagreen}{placeholder source text}}
\\
\\
\textbf{Draft Translation}\\
\textit{\textcolor{darkseagreen}{placeholder draft translation}}
\\
\\
\textbf{Refined Translation}\\
\textit{\textcolor{darkseagreen}{placeholder draft refined translation}}
\\
\\
Please proofread the refined text for grammar, spelling, punctuation, terminology, and overall fluency. Ensure the translation accurately reflects the original meaning and style. Provide only the final, polished translation.
    \end{tabular}}
    \caption{Complete prompts used by the translate step-by-step pipeline.}
    \label{tab:sbys_prompts}
\end{table*}

\begin{table*}[!ht]
    \centering
    \scalebox{0.8}{
    \begin{tabular}{p{1.0\linewidth}}
   \rowcolor{gray!10}
     \hline
    \textbf{\textsc{zero-shot}}\\  
    You are asked to translate the text below into \textcolor{alizarin}{Chinese}. Please output only the translation of the text without any other explanation.\\
     \\
     \textcolor{alizarin}{English}: \textit{\textcolor{darkseagreen}{placeholder source text}}\\
        \textcolor{alizarin}{Chinese}: \\
   \\
   \rowcolor{gray!10}
     \hline
    \textbf{\textsc{zero-shot in context}}\\  
You are asked to translate the text below into \textcolor{alizarin}{Chinese}. You are also given access to the context it appears.\\
\\
Context: \textit{\textcolor{darkseagreen}{placeholder document context}}
\\
\\
Please output only the translation of the text without any other explanation.
\\
\\
\textcolor{alizarin}{English}: \textit{\textcolor{darkseagreen}{placeholder source text}}\\
\textcolor{alizarin}{Chinese}: \\
   \\
\rowcolor{gray!10}
     \hline
    \textbf{\textsc{draft json}}\\   
 Analyze the previous responses and create a JSON object that organizes the linguistic information they contain. This object should have two sections: ``idiomatic$\_$expressions", and ``draft$\_$translation":
\\
\begin{itemize}
    \item \textbf{``idiomatic$\_$expressions"}:
    \begin{itemize}
        \item  This section should also be a list of dictionaries.
        \item Each dictionary represents a phrase and has the keys: ``source$\_$phrase", ``description", ``translation", and ``literal$\_$translation".
        \item The ``translation" key should hold a list of all provided translations for the phrase.
        \item If the response doesn't provide a literal translation, use `null` for the ``literal$\_$translation" value.
        \item If the response doesn't identify relevant idiomatic expressions use `null` for the corresponding value.
        \item Don't include transliterated versions of the translations.
    \end{itemize}
    \item \textbf{``draft$\_$translation"}:
    \begin{itemize}
     \item A string containing the draft translation.
     \item If a translation includes multiple alternatives for the same source text, extract the first one.
     \item Don't include transliterated versions of the translations.
     \item If the translation includes alternatives separated by "$/$", extract the first one.
    \end{itemize}
\end{itemize}

    \end{tabular}}
    \caption{Prompts used for baseline comparisons (zero-shot and zero-shot in context) and structuring the pre-translation research translation artifacts.}
    \label{tab:baseline_prompts}
\end{table*}

\subsection{More example outputs}\label{sec:more_example_outputs}
Table~\ref{tab:example_outputs_comparison} gives more example outputs to support the discussion in \S\ref{sec:qualitative_analysis}.

\begin{table*}[!ht]
    \centering
    \scalebox{0.8}{
    \begin{tabular}{p{0.25\linewidth}p{0.75\linewidth}}
    \toprule
     
\rowcolor{blue!10}
$1$. \bluecheck \textit{Good example} & \\
 \textsc{\textbf{source}} & @user26 cheeked up. I love HER* \\
\textsc{\textbf{reference}} & \zh{@user26 口出狂言。我喜欢她*} \\
& \textit{\textcolor{darkseagreen}{Talking nonsense. I like her*}}\\
\textsc{\textbf{zero-shot}} & \zh{@user26 脸红了。我爱她*} \\
& \textit{\textcolor{darkseagreen}{Blushing. I love her*}}\\
\textsc{\textbf{research and draft}} & \zh{@user26 脸颊做得很饱满。我喜欢她*} \\
& \textit{\textcolor{darkseagreen}{The cheeks are very plump. I like her*}}\\
\textsc{\textbf{refinement}}& \zh{@user26 脸颊做得鼓鼓的，超可爱！我喜欢她*}\\
& \textit{\textcolor{darkseagreen}{The cheeks are very chubby, super cute! I like her*}}\\
\rowcolor{gray!10}
\textbf{\textsc{notes}} & \textit{Both the reference and the zero-shot misinterpreted the phrase ``cheeked up'' in the context of making miniatures. Both the draft and the refined version understood it correctly, with the latter being a bit more colloquial.}\\
\hline
\rowcolor{blue!10}
$2$. \bluecheck \textit{Good example} & \\
 \textsc{\textbf{source}} & I donked up my threading \\
\textsc{\textbf{reference}} & \zh{我完成了这部分学习} \\
& \textit{\textcolor{darkseagreen}{I finished this part of study.}}\\
\textsc{\textbf{zero-shot}} & \zh{我把我的线程搞砸了} \\
& \textit{\textcolor{darkseagreen}{I messed up my threads [note: a terminology in computing].}}\\
\textsc{\textbf{research and draft}} & \zh{我搞砸了我的帖子顺序} \\
& \textit{\textcolor{darkseagreen}{I messed up the order of my posts.}}\\
\textsc{\textbf{refinement}}& \zh{我搞砸了帖子顺序}\\
& \textit{\textcolor{darkseagreen}{I messed up the post order.}}\\
\rowcolor{gray!10}
\textbf{\textsc{notes}} & \textit{The reference misinterpreted both ``donked up'' and ``threading'' in this context, and the zero-shot translation mistook the latter for the computing terminology.}\\
\hline
\rowcolor{blue!10}
$2$. \bluecheck \textit{Good example} & \\
 \textsc{\textbf{source}} & It is a ridiculous level.
  \\
\textsc{\textbf{reference}} & \zh{这关太难了。} \\
& \textit{\textcolor{darkseagreen}{This level is too hard.}}\\
\textsc{\textbf{zero-shot}} & \zh{这是一个荒谬的关卡。} \\
& \textit{\textcolor{darkseagreen}{This is a ridiculous level.}}\\
\textsc{\textbf{research and draft}} & \zh{这是一个荒谬的关卡。} \\
& \textit{\textcolor{darkseagreen}{This is a ridiculous level.}}\\
\textsc{\textbf{refinement}}& \zh{这个关卡真是太难了。}\\
& \textit{\textcolor{darkseagreen}{This level is really too hard.}}\\
\rowcolor{gray!10}
\textbf{\textsc{notes}} & \textit{Zero-shot and draft translations are direct, literal, and stylistically awkward in Chinese, especially in the social media domain.}\\
\hline
\rowcolor{red!10}
$4$. \xmark \ \ \textit{Bad example} & \\
 \textsc{\textbf{source}} & Ivory chuckled from the memory.  \\
\textsc{\textbf{reference}} & \zh{艾弗里回忆到这里，笑了起来。} \\
& \textit{\textcolor{alizarin}{Ivory thought of this and laughed}}\\
\textsc{\textbf{zero-shot}} & \zh{艾弗里想起这件事，不禁轻笑起来。} \\
& \textit{\textcolor{alizarin}{Ivory thought of this and couldn't help starting to chuckle.}}\\
\textsc{\textbf{research and draft}} & \zh{艾弗里回想起当时的情景，不禁轻笑出声} \\
& \textit{\textcolor{alizarin}{Ivory thought of the situation at that time and couldn't help letting out a chuckle.}}\\
\textsc{\textbf{refinement}}& \zh{艾弗里摸了摸肩膀上上次留下的瘀伤，不禁轻笑出声。}\\
& \textit{\textcolor{alizarin}{Ivory touched the bruise on the shoulder left from the last time and couldn't help letting out a chuckle.}}\\
\rowcolor{gray!10}
\textbf{\textsc{notes}} & \textit{All translations did a good job translating the phrase ``chuckled from the memory'' fluently into Chinese, but the refinement hallucinated by adding ``touched the bruise.''}\\
\hline
\rowcolor{red!10}
$5$. \xmark \ \ \textit{Bad example} & \\
 \textsc{\textbf{source}} & one of them ran for something and grabbed it \\
\textsc{\textbf{reference}} & \zh{其中一个人跑过去抓起了什么东西} \\
& \textit{\textcolor{alizarin}{One of them ran over and grabbed something.}}\\
\textsc{\textbf{zero-shot}} & \zh{其中一个人跑去拿了什么东西} \\
& \textit{\textcolor{alizarin}{One of them ran and grabbed something}}\\
\textsc{\textbf{research and draft}} & \zh{其中一个人跑到某个东西那里，抓起它} \\
& \textit{\textcolor{alizarin}{One of them ran to something and grabbed it.}}\\
\textsc{\textbf{refinement}}& \zh{其中一个人跑到某个东西那里，抓起它}\\
& \textit{\textcolor{alizarin}{One of them ran to something and grabbed it.}}\\
\rowcolor{gray!10}
\textbf{\textsc{notes}} & \textit{Although all translations are adequate and capture the semantic meaning of the source correctly, the draft and the refinement keep the original source structure and the resulting translation sounds like
translationese.}\\
\bottomrule
    \end{tabular}}
    \caption{Sample of step-by-step and zero-shot outputs along with notes. }
    \label{tab:example_outputs_comparison}
\end{table*}

\label{sec:appendix}

\end{document}